\documentclass{article}
\usepackage{ijcai22}

\usepackage{amsfonts} 
\usepackage{color,soul}
\usepackage{times}
\usepackage{soul}
\usepackage{url}
\usepackage[hidelinks]{hyperref}
\usepackage[utf8]{inputenc}
\usepackage[small]{caption}
\usepackage{graphicx}
\usepackage{amsmath}
\usepackage{amsthm}
\usepackage{booktabs}
\usepackage{algorithm}
\usepackage{algorithmic}
\usepackage{subfigure}
\title{CGMN: A Contrastive Graph Matching Network for Self-Supervised Graph Similarity Learning}

\author{
    Di Jin$^1$\and Luzhi Wang$^1$\and Yizhen Zheng$^2$\and Xiang Li$^3$\and Fei Jiang$^3$\and Wei Lin$^3$\And Shirui Pan$^2$\footnote{Corresponding author}\\
    \affiliations
    $^1$College of Intelligence and Computing, Tianjin University, Tianjin, China\\
    $^2$Department of Data Science and AI, Faculty of IT, Monash University, Australia\\
    $^3$Meituan, Beijing, China\\
    \emails
    \{jindi, wangluzhi\}@tju.edu.cn,
    \{yizhen.zheng1, shirui.pan\}@monash.edu
}

\begin{document}

\maketitle

\begin{abstract}
Graph similarity learning refers to calculating the similarity score between two graphs, which is required in many realistic applications, such as visual tracking, graph classification, and collaborative filtering. As most of the existing graph neural networks yield effective graph representations of a single graph, little effort has been made for jointly learning two graph representations and calculating their similarity score. In addition, existing unsupervised graph similarity learning methods are mainly clustering-based, which ignores the valuable information embodied in graph pairs. To this end, we propose a contrastive graph matching network (CGMN) for self-supervised graph similarity learning in order to calculate the similarity between any two input graph objects. Specifically, we generate two augmented views for each graph in a pair respectively. Then, we employ two strategies, namely cross-view interaction and cross-graph interaction, for effective node representation learning. The former is resorted to strengthen the consistency of node representations in two views. The latter is utilized to identify node differences between different graphs. Finally, we transform node representations into graph-level representations via pooling operations for graph similarity computation. We have evaluated CGMN on eight real-world datasets, and the experiment results show that the proposed new approach is superior to the state-of-the-art methods in graph similarity learning downstream tasks.
\end{abstract}

\section{Introduction}
\begin{figure}[t]
	\centering
	\includegraphics[width=0.45\textwidth]{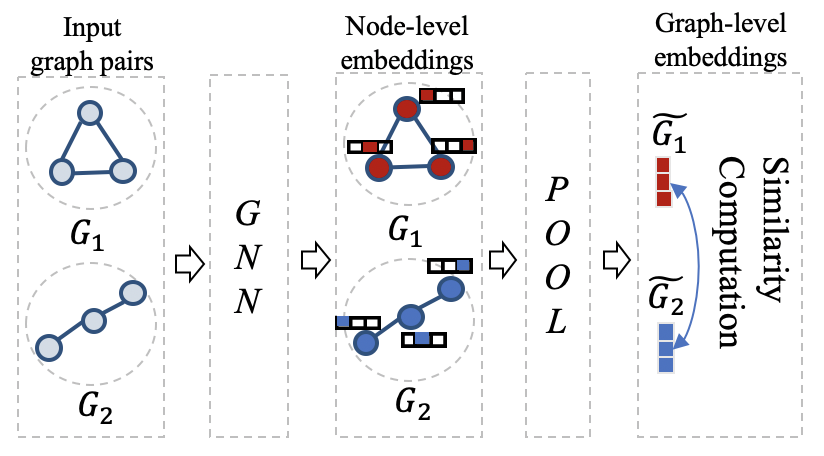} 
	\caption{A view of the supervised graph similarity learning model.}
	\label{f1}
\end{figure}

Graph-structured data is a natural representation in coding structures that exist in a variety of domains, including computer vision \cite{wan2021dual}, social network \cite{jin2021heterogeneous}, NLP \cite{yu2021gcn}, recommender systems \cite{jia2021hypergraph,jin2021survey}, etc. Graph similarity learning is one of the most important research problems in exploring graph-structured data. It aims to learn a similarity score for a pair of input graphs. Graph similarity learning has a wide range of applications in various scenarios, such as binary code analysis \cite{xu2017neural}, collaborative filtering algorithms for recommender systems \cite{DBLP:conf/sigir/SuZEG21}, etc. Different task scenarios have different similarity metrics, e.g., graph edit distance (GED) \cite{sanfeliu1983distance}, and maximum common subgraph (MCS) \cite{bunke1997relation}. The GED problem refers to solving the minimum number of editing operations required to convert one graph to the other, and MCS aims to find the largest induced subgraph common to two given graphs. However, GED and MCS problems are a class of NP-complete problems \cite{wang2021combinatorial,bai2021glsearch}, indicating that solving such problems is computationally expensive. Some traditional methods, such as A* \cite{neuhaus2006fast} and Hungarian \cite{riesen2009approximate}, can accurately solve the problem of GED, but they have very high computational complexity and thus are hard to be applied on large-scale real datasets. In addition, traditional search algorithms mainly consider the graph structure, while ignoring the rich attribute information contained in the graph, including node features, edge features, etc, limiting their applicability. 

Given the great difficulty of computing the graph distances, people begin to use graph neural networks (GNNs) to learn the similarity between two graphs \cite{li2019graph}. 
But most of the current GNN-based graph similarity learning approaches are supervised and normally adopt a training scheme, which can be summarized as follows. First, the representation of nodes in each graph are learned by using a GNN encoder. Then, graph-level representations are obtained by using a pooling operation. After that, with a distance metric (e.g., cosine similarity and Euclidean distance), the similarity score can be obtained given a pair of graph-level representations. Finally, with the guidance of ground truth, they train the model with the computed loss, which is the difference between the similarity score and the ground truth labels. Fig. \ref{f1} shows the details of this process. However, the heavy reliance of these methods on labels hinders them from being applied in real-world applications where the labeling information is usually scarce and expensive to obtain. To address this problem, we opt to design an unsupervised deep graph similarity learning algorithm.

\begin{figure}[t]
	\centering
	\includegraphics[width=0.40\textwidth]{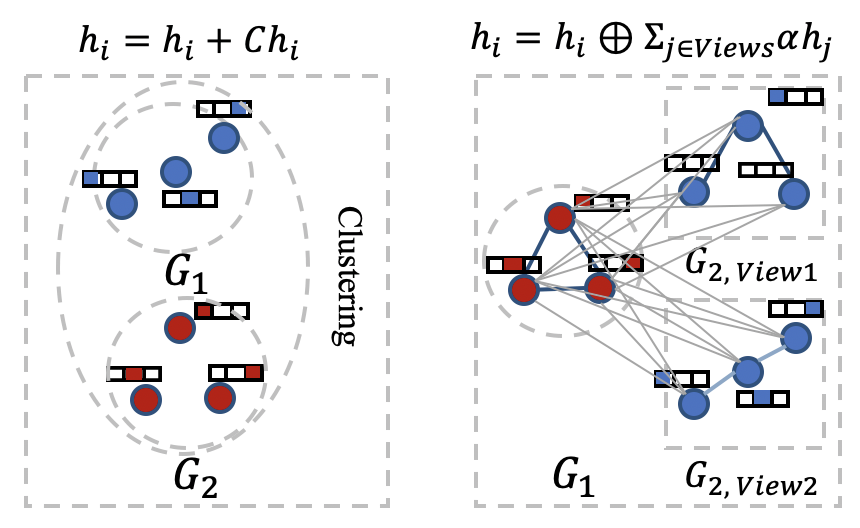} 
	\caption{Comparison of node information update process. Left: Clustering-based unsupervised graph similarity learning. Right: Our proposed model CGMN. $h_i$ denotes the embedding of node $i$, $C$ represents the clustering weight, $\oplus$ is concatenation, and $\alpha$ represents the cross-graph interaction weight.}
	\label{f2}
\end{figure}

However, the current unsupervised GNN methods mainly focus on node (graph) classification or clustering tasks, and limited works have explored graph similarity learning. There are a few unsupervised graph similarity learning studies, such as GA-MGMC \cite{wang2020graduated} and ScGSLC \cite{li2021scgslc}. 
The main idea of these works can be summarized as follows.
First, these works cluster all graphs and compute the clustering weight $C$ for any two graphs, where $C$ is learned based on the clustering results (i.e., graphs in the same cluster have higher $C$, and vice versa). 
Then, node embeddings are updated with $C$ and pooled to generate the graph-level embeddings. And finally, these works use the pseudo labels and the learned similarity score to calculate the loss. Fig. \ref{f2} (left) briefly describes the process of updating node information. 
These clustering-based methods suffer from limitations including the negligence of rich information in cross-graph interaction, as well as the manual choice of the number of clusters. Specifically, they rarely consider the node-to-node matching across graphs, which can provide abundant supervision signals for model training \cite{li2019graph}. Furthermore, they require to manually choose an optimal number of clusters for training, which is essential in clustering but rather difficult to set in practice. Thus, these methods typically lead to suboptimal solutions.

In order to solve the aforementioned problems, we design a novel scheme for unsupervised graph similarity learning, named as \textbf{C}ontrastive \textbf{G}raph \textbf{M}atching \textbf{N}etwork (CGMN). We construct a node embedding method based on augmented views and aim to compute graph similarity scores between graphs. We first augment two different views for each graph in a graph pair to provide different contexts. With a designed cross-view interaction module, for a target graph, we weightily aggregate all node embeddings in the other view to obtain the view-level embedding. After that, we conduct cross-graph matching, which outputs a graph-level embedding for the target node, to distill the rich information embodied in cross-graph interaction. Both the view- and graph-level embedding are concatenated to the node embedding, which is used in building contrastiveness between two views to learn node embeddings of the graph.
 
We expect cross-graph matching of nodes to occur during the learning process, so we interact node information between one graph and views of the other graph, instead of interacting at the end of node representation training. Interacting nodes with different contexts shows the node-to-node similarity from different views. After the node embedding training, we obtain the representation of the whole graph by the pooling module. Similarity scores are calculated for different tasks using graph-level embedding. Fig. \ref{f2} (right) illustrates the process of node information update.

Our contributions can be summarized as follows:
\begin{itemize}
\item We introduce contrastive learning to the unsupervised graph similarity model, which improves the accuracy of graph similarity through node-to-node matching instead of generally using clustering to simulate similarity calculation. 
\item We propose \textit{cross-view interactions} and \textit{cross-graph interactions} in unsupervised graph similarity learning in the same framework, which enhances agreement of nodes and fills the gap that two graphs cannot interact with each other in clustering-based methods. 
\item We conduct experiments on eight real-world datasets, and the experimental results validate the superiority of our proposed new method compared with SOTA.

\end{itemize}

\section{Related Work}

\subsection{Graph Similarity Learning}
Inspired by recent advances in deep learning, computing graph similarity with deep networks has received increasing attention. The first category is supervised graph similarity learning, which is a line of work that uses deep feature encoders to learn the similarity of the input pair of graphs. The learned similarity scores and ground truth are fed into the loss function to optimize the encoder with resulting in an end-to-end learning scheme. For example, GMN \cite{li2019graph}, SimGNN \cite{bai2019simgnn}, and H2MN \cite{zhang2021h2mn} all belong to this category, which use GNNs as their encoder. The second category is the unsupervised graph similarity learning, such as GA-MGMC \cite{wang2020graduated}, ScGSLC \cite{li2021scgslc}, etc. The process of these works stands to encode the nodes using a deep encoder and update the features of the nodes by using the clustering weights. But they are limited to clustering and thus difficult to learn the cross-graph interaction information of two graphs.

\subsection{Graph Contrastive Learning}

Contrastive learning is popular in self-supervised graph representation learning, which aims to learn discriminative representations by comparing positive and negative samples~\cite{he2020momentum,jin2021multi}. 
For example, Deep Graph Infomax (DGI) \cite{velickovic2019deep} extends Deep Infomax \cite{DBLP:conf/iclr/HjelmFLGBTB19} by comparing nodes and graphs to learn node representations. 
MVGRL \cite{hassani2020contrastive} introduces graph diffusion to graph augmentation by comparing different graph structures to learn node representations. 
GRACE \cite{grace} maximizes the agreement of node representations between two augmented graph views.
Contrastive learning has shown impressive results in unsupervised graph representation learning, but it is typically used for classification and clustering tasks. There are still no models designed for graph similarity learning tasks as far as we know.

\begin{figure*}[tb]
	\centering
	\includegraphics[width=0.95\textwidth]{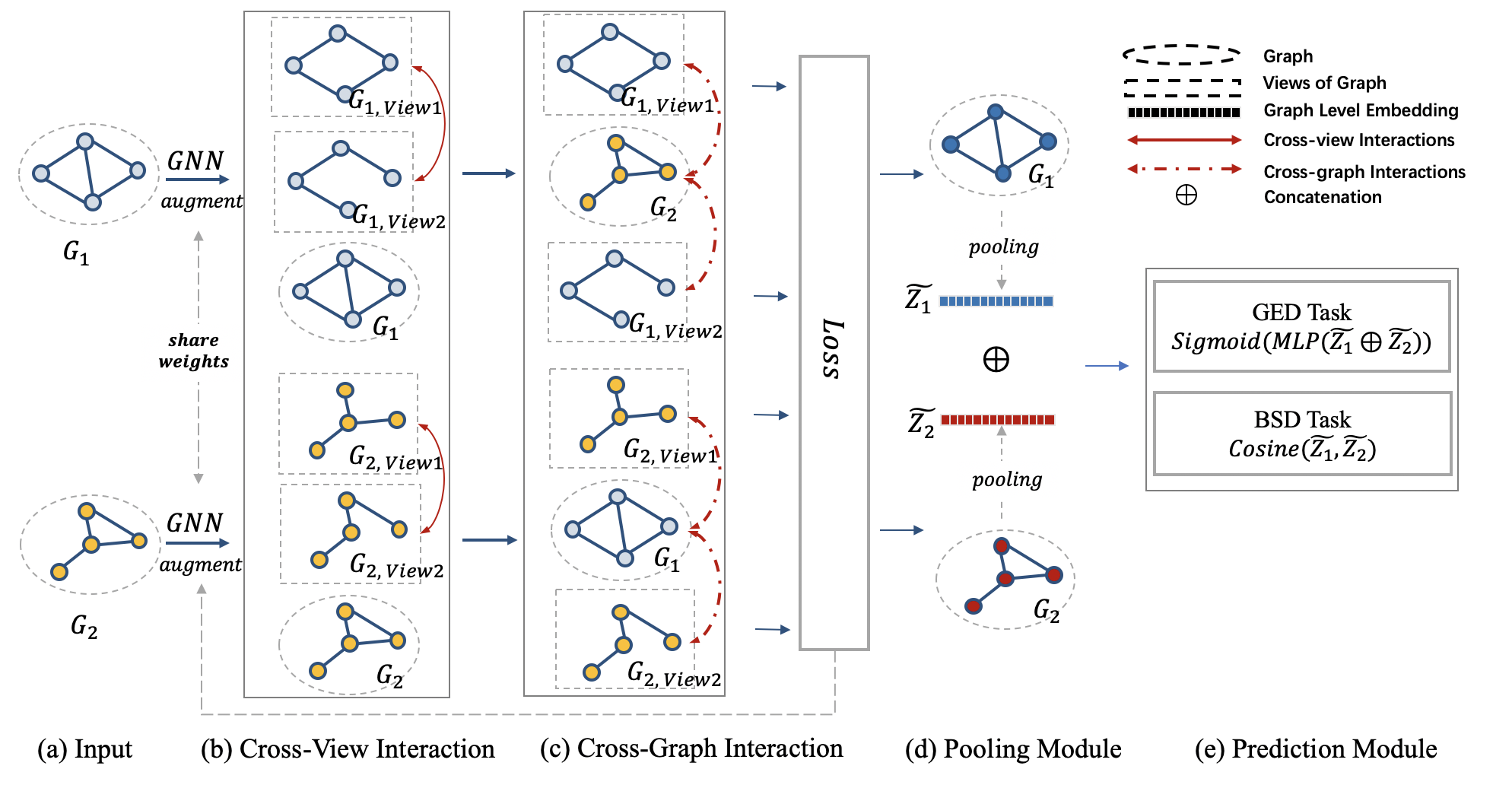} 
	\caption{Overview of CGMN. First, we provide a framework to learn the embedding of each node. Second, we propose a cross-graph interaction strategy to match nodes in graph pairs. Third, we aggregate node embeddings to obtain the graph-level representations. Finally, we predict the similarity of graphs for different tasks.}
	\label{flowchart}
\end{figure*}

\section{Method}

\subsection{Problem Definition}
An undirected attributed graph $G={(V,E,A,X)}$ consists of a set of nodes $V$ with $|V|=n$, a set of edges $E$ with $|E|=m$, the adjacency matrix $A \in \mathbb{R}^{n\times n}$ and node attribute matrix $X\in \mathbb{R}^{n\times d}$ with feature dimension $d$. If there is an edge between node $i$ and node $j$, $A_{i,j} = 1$, else $A_{i,j}=0$. For each node, its features are represented as a vector $x\in \mathbb{R}^d$, where $x_i$ denotes feature values of node $i$. Therefore, $X\in \mathbb{R}^{n\times d}$ denotes the node attribute matrix of the graph and each node has $d$ features. 
Given two graphs $G_1$ and $G_2$ as input, the graph similarity learning is to calculate a similarity score $y$ in order to measure the difference between two graphs in an input graph pair. Different similarity metrics can be defined according to different downstream tasks. Our method tries to learn an encoder to generate graph-level embeddings of two graphs and calculate their similarity score.

\subsection{Method Overview}
Fig. \ref{flowchart} shows our overall framework CGMN. Firstly, in part (a), two graphs $G_1$ and $G_2$ are fed into CGMN. Then, CGMN generates two correlated graph views via graph augmentation techniques for $G_1$ and $G_2$ respectively. After that, the original input graphs and their views are embedded into a low-dimensional vector space by a GNN encoder. Secondly, a cross-view interaction strategy (b) and a cross-graph interaction strategy (c) are provided to learn effective node embeddings. In part (d), graph-level representations are obtained by pooling the node embeddings. Finally, CGMN computes similarity using two graph-level representations for the subsequent prediction tasks (e). Details will be introduced in the rest of this section.

\paragraph{Node embedding layer.}
The siamese network consists of two identical sub-networks and excels at comparing the subtle differences between the two inputs. We integrate a siamese network architecture with contrastive learning-based GNN as the basic node embedding layer of CGMN. Graph augmentation, a key component of graph contrastive learning, is used to generate different views to provide diverse contexts for nodes, which facilitates the optimization of contrastive objectives.
We generate two correlated graph views by randomly performing corruption techniques, such as masking node features and removing edges. Positive and negative samples are generated to learn node representations discriminatively by comparing them. 
After graph augmentation, we consider using a multi-layer GCN to generate node embeddings of the input graphs and their views:
\begin{equation}
\begin{aligned}
H^l  = \sigma (\widetilde{A} H^{l-1} W^{l-1}),
\end{aligned}
\label{eq: intra1}
\end{equation}
where $H$ is the set of node embeddings of a graph (view), $l$ is the number of layers, $\sigma$ is the activation function, $\widetilde{A}$ is the normalized Laplacian matrix, and $W$ is the layer-specific trainable weighted matrix.

\paragraph{Cross-view interaction.}
Inspired by works on graph contrastive learning, the agreement of corresponding nodes in the two views should be maximized. Specifically, if node $u\in G_{1, View1}$ corresponds to node $v \in G_{1, View2}$, node $u$'s embedding $h_u$ and node $v$'s embedding $h_v$ can be regarded as a pair of positive samples, and embeddings of other nodes in the two views can be naturally regarded as negative samples of $h_u$ \cite{grace}. 
To maximize agreement between node embeddings, we need to expand the similarity between positive samples and reduce the similarity between negative samples. The similarity of two nodes' embeddings in vector space can be defined as: 
\begin{equation}
\begin{aligned}
    \texttt{sim}(h_u, h_v) := \texttt{exp}(\texttt{cos}\left(h_u, h_v \right) / \tau),
\end{aligned}
\end{equation}
here, \texttt{cos}($\cdot$) donates the cosine similarity function, \texttt{exp}($\cdot$) is the exponential function, and $\tau$ is a temperature parameter to help the model learn from hard negatives which are closest in distance, but farther than positive examples \cite{simclr}.

\begin{figure}[ht]
	\centering
	\includegraphics[width=0.35\textwidth]{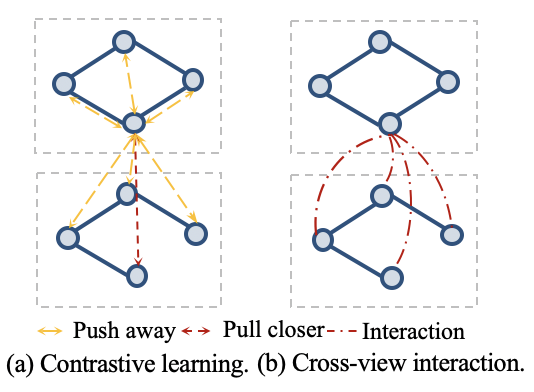} 
	\caption{Difference between (a) contrastive learning and (b) cross-view interaction.}
	\label{crossview}
\end{figure}

However, the aforementioned processes mainly consider the corresponding nodes' agreement between views. The similarity between the nodes in the two views has not been fully considered. To address this drawback, we propose a cross-view interaction method that measures node-to-node similarity with cosine similarity, which enriches the self-supervised signals.

\begin{equation}
\begin{aligned}
\hat{h_u} &= h_u \oplus \sum_{v\in G_{1, View2}}\texttt{cos}(h_u,h_v)h_v,
\end{aligned}
\end{equation}
where $\oplus$ is the concatenation. Fig. \ref{crossview} shows the difference the contrastive learning and cross-view interaction.

Cross-view interactions help maintain the monotonic increasing correlation between node similarity and node agreement. If nodes in the two views are corresponding, they have higher cross-view interaction scores than others, which means that their embeddings are close in the vector space and promote the maximization of node agreements. 

\paragraph{Cross-graph interaction.}
Learning node-to-node matching between two graphs is a \textit{key} part of graph similarity learning. If we perform cross-graph matching after the node representation training is completed, we may not be able to better learn the interaction information of the two graphs. In the above process, we augment each graph with two views, which can replace the graph to participate in the cross-graph information interaction during the training process. 
Specifically, we interact the original graph $G_1$ with each view of $G_2$. We use the attention mechanism to implement cross-graph information interaction. We design the attention function for computing the similarity score between two node embedding vectors. We use cosine similarity as a concrete form of attention mechanism. Assuming that node $u$ belongs to the graph $G_1$, then node $u$'s representation with cross-graph information can be calculated as:
\begin{equation}
\begin{aligned}
h_u^* = h_u &\oplus \sum_{i\in G_{2, View1}}\texttt{cos}(h_u,\hat{h_i})\hat{h_i} \\
 &\oplus \sum_{j\in G_{2, View2}}\texttt{cos}(h_u,\hat{h_j})\hat{h_j}.
\end{aligned}
\end{equation}

Cross-graph interaction and cross-view interaction are different. The main difference lies in not only the object of action, but also the purpose. The purpose of cross-view interaction is to maximize node agreement in different views, whereas cross-graph interaction aims to learn the node matching between different graphs. After cross-view interaction and cross-graph interaction learning, we construct the self-supervised loss by maximizing the agreement of positive samples and minimizing the agreement of negative samples:
\begin{equation}
    loss(h_u^*,h_v^*) = -\log\frac{\texttt{sim}(h_u^*,h_v^*)}{ \texttt{sim}(h_u^*,h_v^*)+\sum_{k=1}^{N}\texttt{sim}(h_u^*,h_k^*)}, 
\end{equation}
where $u$ and $v$ are a pair of positive samples and $N$ is the number of negative samples. Both node $u$ and node $v$ need to be calculated, then the total loss can be defined as their average result:
\begin{equation}
    \mathcal{L}=\frac{1}{2}\left[ loss(h_u^*,h_v^*)+loss(h_v^*,h_u^*)\right].
\end{equation}

\paragraph{Prediction layer.}
We calculate graph similarity by graph-level embeddings. For simplicity, we employ mean pooling as the default aggregation function:
\begin{equation}
    \Tilde{Z} = \texttt{AVG}(h_u: u\in G),
\end{equation}
where \texttt{AVG}($\cdot$) denotes the average function.

For different prediction tasks, we can define different methods for calculating similarity scores. In \textbf{GED} task, given two graphs $G_1$ and $G_2$ as input, $GED(G_1,G_2)$ records the number of edits that transform $G_1$ to $G_2$. An edit can be defined as an insertion or a deletion operation of a node or edge. 
The number of GED is converted to consecutive values in $(0, 1]$ by $exp(-\frac{GED(G_1,G_2)}{|G_1|+|G_2|})$. In our implementation, we use a multi-layer perceptron (MLP) to obtain the GED score, which takes the concatenation of the two graph-level embeddings as input. The similarity score is then calculated as:
\begin{equation}
     y = \texttt{sigmoid}(\texttt{MLP}(\Tilde{Z_1} \oplus \Tilde{Z_2})).
\end{equation}

The \textbf{binary similarity detection (BSD)} task aims to predict whether the given graphs are similar or not. The input two graphs are assigned a label $\hat{y}=\{1,-1\}$ where $1$ indicates the input two graphs are similar, and $-1$ indicates these two graphs are not similar. Our goal is to learn the similarity between the input two graphs, and the calculated similarity score will be classified into similar ($1$) or dissimilar ($-1$). We use the following cosine similarity as the similarity metric:
\begin{equation}
     y = \texttt{cos}(\Tilde{Z_1},\Tilde{Z_2}).
\end{equation}

We evaluate the model performance by comparing predicted similarity scores with ground truth, and detail the performance of our proposed model on these two types of tasks in subsequent sections.

\section{Experiments}
In this section, we systematically evaluate our proposed model on GED and BSD tasks with eight benchmark datasets. Specifically, we first introduce the datasets for these two mentioned tasks, and then describe the baselines that are to be compared with our model, and the details of the experimental setup. We finally present the main evaluation results and perform the ablation experiment.

\subsection{Datasets}%
We evaluate our model on eight public datasets, which are commonly used for graph similarity problem \cite{9516695}. The statistics of datasets are summarized in Table \ref{dataset}. In the GED task, we employ two benchmark datasets, including Aids700nef and Linux1000. We followed the previous works \cite{9516695} and divide each dataset into $60\%$, $20\%$, $20\%$, for training, validation, and testing set respectively.
OpenSSL (OS) and FFmpeg (FF) datasets are employed for the BSD task, where [3, 200] means the size ranges of pairs of graphs. The datasets are divided into $10\%$, $10\%$, $80\%$ for training/ validation/ testing set for supervised methods and $80\%$, $10\%$, $10\%$ for unsupervised methods.

\subsection{Baselines $\&$ Settings}
We compare CGMN with two supervised GNN methods and three state-of-the-art unsupervised GNN models. Supervised methods include GCN \cite{DBLP:conf/iclr/KipfW17} and GIN \cite{DBLP:conf/iclr/XuHLJ19}. Unsupervised methods include DGI \cite{velickovic2019deep}, GRACE \cite{grace}, and ScGSLC \cite{li2021scgslc}. The first four are graph representation learning methods, and the last one is an unsupervised clustering-based graph similarity learning method.

\begin{table}[tbph]
\small
\centering
\renewcommand\tabcolsep{10pt}
\begin{tabular}
{@{}lcccc@{}}
	\toprule
		\textbf{Datasets} & \textbf{Graphs} & \textbf{ AvgN} & \textbf{AvgE} & \textbf{Classes}  \\
		\midrule
				Aids700nef     &700   & 8.90          & 8.80        & -               \\
		Linux1000     &1000   & 7.58         & 6.94       & -               \\
		\midrule
	    OS [3, 200]    &73,953  & 15.73           & 21.97              & 4,249            \\
	    
		OS [20, 200]   &15,800  &44.89            & 67.15             & 1,073            \\
		OS [50, 200]   &4,308   & 83.68           & 127.75              & 338               \\
		FF [3, 200]      &83,008  &18.83          & 27.02               & 10,376                \\
		
		FF [20, 200]     &31,696  & 51.02            & 75.88             & 7,668               \\
		FF [50, 200]     &10,824   & 90.93          & 136.83        & 3,178               \\
		 \bottomrule

	\end{tabular}
\caption{Statistics of the datasets.}
	\label{dataset}
\end{table}
We employ a 3-layers GCN as our core encoder. The learning rate is uniformly set as $0.0001$ and the latent dimension of graph encoders, projectors, and the predictor are fixed to $100$. We regard the GED task and BSD task as downstream tasks. In the GED task, we use the same settings for the supervised baselines, including dataset partitioning, learning rate, etc. We train the supervised models using mean square error (MSE) loss with all the ground truth from the training set. For unsupervised baselines, we use a multi-layer perceptron (MLP) to map the predicted similarity scores to GED scores distribution intervals, for which we fine-tune the mapping between the predicted scores and the GED scores by using a MSE loss with $1\%$ GED ground truth. In the BSD task, we train supervised models using the same approaches as GED. For unsupervised baselines, we use their self-contained losses for graph representation learning. We repeat all experiments ten times and report the average score and standard deviation.

\begin{table*}[htpb]
\small
\centering
\renewcommand\tabcolsep{9.4pt}
\begin{tabular}
{clccccc} \toprule
 \textbf{Datasets}          & \textbf{Methods}  &\textbf{ MSE ($10^{-3}$)} & \textbf{$\rho$}              & \textbf{$\tau$}            & \textbf{p@10}            & \textbf{p@20 }           \\ \midrule 
           & GCN    & 11.395$\pm$1.315               & 0.577$\pm$0.021     & 0.418$\pm$0.018                & 0.041$\pm$0.002 & 0.077$\pm$0.003 \\ 
           & GIN    & 9.280$\pm$0.163                 & 0.629$\pm$0.020     & 0.462$\pm$0.016                & 0.044$\pm$0.018 & 0.096$\pm$0.021 \\ 
Aids700nef & DGI    & 15.009$\pm$0.347                & 0.231$\pm$0.093     & 0.164$\pm$0.061                & 0.039$\pm$0.006 & 0.076$\pm$0.001 \\ 
           & GRACE  & 12.176$\pm$1.693                & 0.366$\pm$0.186     & 0.261$\pm$0.134                & 0.038$\pm$0.004 & 0.072$\pm$0.018 \\ 
           & ScGSLC & 13.060$\pm$0.193                & 0.394$\pm$0.133     & 0.281$\pm$0.097                & 0.080$\pm$0.026 & \textbf{0.142$\pm$0.044 }\\ 
           & CGMN  & \textbf{6.641$\pm$2.227}                & \textbf{0.674$\pm$0.129  }    & \textbf{0.502$\pm$0.107}               & \textbf{0.084$\pm$0.019} & 0.140$\pm$0.024 \\ \midrule 
           
           & GCN    & 11.986$\pm$1.532                & 0.569$\pm$0.033     & 0.411$\pm$0.028 & 0.043$\pm$0.005 & 0.071$\pm$0.001 \\ 
           & GIN    & 22.188$\pm$5.259                & 0.647$\pm$0.112     & 0.484$\pm$0.099                & 0.081$\pm$0.018 & 0.084$\pm$0.025 \\ 
Linux1000      & DGI    & 33.854$\pm$0.013                & 0.052$\pm$0.018     & 0.039$\pm$0.002                & 0.035$\pm$0.020 & 0.073$\pm$0.016 \\ 
           & GRACE  & 14.180$\pm$2.080                & 0.852$\pm$0.019     & 0.673$\pm$0.025                & \textbf{0.443$\pm$0.155} & \textbf{0.452$\pm$0.175} \\ 
           & ScGSLC & 13.423$\pm$2.038                & 0.840$\pm$0.010     & 0.658$\pm$0.021                & 0.192$\pm$0.095 & 0.213$\pm$0.120 \\ 
           & CGMN  & \textbf{10.514$\pm$1.178}                & \textbf{0.873$\pm$0.013}     & \textbf{0.700$\pm$0.015}                & 0.307$\pm$0.071 & 0.330$\pm$0.091 \\ 
 \bottomrule
\end{tabular}
\caption{Experimental results on the GED datasets in terms of five evaluation metrics.}
\label{ged}

\end{table*}

\begin{table*}[htpb]
\centering
\small
\renewcommand\tabcolsep{10pt}
\begin{tabular}
{lllcccc} \toprule
\textbf{Methods} & \textbf{OS [50, 200]} & \textbf{OS [20, 200]} & \textbf{OS [3, 200]} & \textbf{FF [50, 200]}& \textbf{FF [20, 200]} & \textbf{FF [3, 200]}\\ 
\midrule

GCN  & 67.24$\pm$1.14    & 68.09$\pm$1.01     & 73.51$\pm$0.72     & 78.41$\pm$0.49     & 79.47$\pm$0.08    & 80.88$\pm$0.18  \\ 
GIN  & 66.60$\pm$0.10     & 63.85$\pm$0.56   &  75.65$\pm$0.30    & 78.38$\pm$0.20    & 81.25$\pm$0.57      &\textbf{81.82$\pm$0.25 } \\

DGI  & 67.55$\pm$2.76    & 63.58$\pm$1.96    & 72.58$\pm$2.36    & 86.10$\pm$0.66  & 80.82$\pm$2.22     &66.28$\pm$0.30 \\
GRACE  & 68.84$\pm$2.45 & 67.01$\pm$0.49  & 69.86$\pm$0.29     &85.44$\pm$0.27    & 75.05$\pm$0.73    &66.95$\pm$2.78 \\

ScGSLC & 67.43$\pm$0.82   & 61.46$\pm$0.33   & 63.28$\pm$0.09    & \textbf{87.57$\pm$0.82}  & 83.27$\pm$0.71   & 69.80$\pm$1.22 \\ 

CGMN  & \textbf{80.89$\pm$0.20}   & \textbf{78.15$\pm$0.85}  &\textbf{75.94$\pm$1.86}   & 86.11$\pm$0.98   & \textbf{86.76$\pm$0.85 }   &77.98$\pm$2.69 \\ 
		
 \bottomrule
\end{tabular}
\caption{Experimental results on the BSD datasets in terms of AUC scores (\%).}
\label{cfg}
\end{table*}

\subsection{Evaluation Results}
We systematically evaluate the overall performance on the two tasks, the ablation study, and the parameter analysis. 

\paragraph{GED task.}
We mainly use MSE as an evaluation metric of the GED task, which measures the mean squared error between the predicted similarity scores and the GED ground truth. We also provide other evaluation metrics, such as Spearman's Rank Correlation Coefficient ($\rho$) and Kendall's Rank Correlation Coefficient ($\tau$) to comprehensively measure how well the predicted scores match the true scores. Precision at $k$ ($p@k$) is calculated by dividing the intersection of the predicted top $k$ results and the ground truth top $k$ results. 
As shown in Table \ref{ged}, our model CGMN outperforms other baseline methods in four metrics on Aids700nef and even partially surpasses the supervised algorithm. Compared to the optimal supervised models, the MSE value of CGMN has been increased by $28.43\%$, and compared with the unsupervised algorithms, the MSE value increases by $45.46\%$. On Linux1000 dataset, we achieve optimality in three metrics and suboptimality in the other two evaluation metrics. Compared with the supervised models and the unsupervised models, our model CGMN improves the value of MSE by $12.28\%$ and $21.67\%$ respectively, which demonstrates the superiority of our new model.

\paragraph{BSD task.}
We use the area under the ROC curve (AUC) as an evaluation metric for the BSD task, where AUC and model performance are positively correlated. The overall results are presented in Table \ref{cfg}. We bold the optimal results and `$\pm$' indicates a numerical range. Compared to unsupervised methods, our model CGMN consistently achieves the superior performance in terms of all metrics on all six datasets. CGMN can learn a good embedding function, which can be generalized to invisible test plots.

\begin{table}[t]
\centering
\small
\tabcolsep 3.5pt
\renewcommand\arraystretch{1.1}
\begin{tabular}
{clcccc} \toprule
 \textbf{Methods}  &\textbf{MSE} & \textbf{$\rho$}              & \textbf{$\tau$}            & \textbf{p@10}            & \textbf{p@20 }  \\ \midrule
CGMN $w/o$ cross-view  & 8.239 & 0.614 & 0.451 & 0.064  & 0.114  \\
CGMN $w/o$ cross-graph & 8.753 & 0.537 & 0.387 & 0.050  & 0.091  \\
CGMN                 & 6.641 & 0.674 & 0.502 & 0.084  & 0.140 \\  \bottomrule
\end{tabular}
\caption{Ablation study on Aids700nef.}
\label{abs}
\end{table}

\paragraph{Ablation study.}
To verify the effectiveness of cross-view interaction and cross-graph interaction, we employ two CGMN variants on Aids700nef, each of which has one of the key components removed. 
Table \ref{abs} shows the results of the ablation study, where $w/o$ represents the removal of a strategy. 
Experimental results show that the performance of our model degrades without one of the key strategies on Aids700nef. We find that the CGMN achieves $24.13\%$ higher relative to the others variants on MSE ($10^{-3}$). Among other evaluation metrics, CGMN also achieves the best performance compared to its two variants, which proves the effectiveness of the combination of our two proposed schemes.

\begin{figure}
	\centering
	\small
	\subfigure[{Masking node features.}]{\includegraphics[width=0.44\columnwidth]{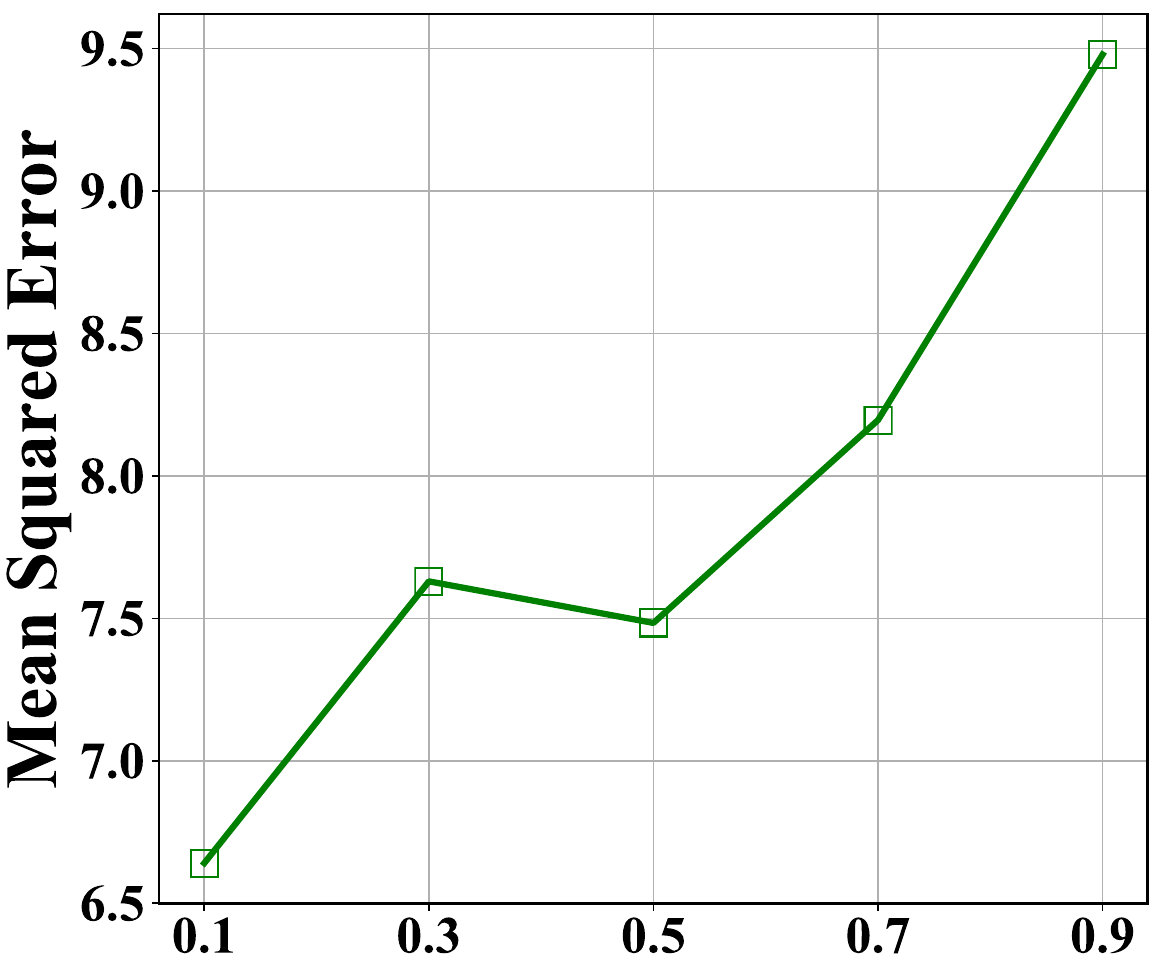}}
	\subfigure[{Removing edges.}]{\includegraphics[width=0.4\columnwidth]{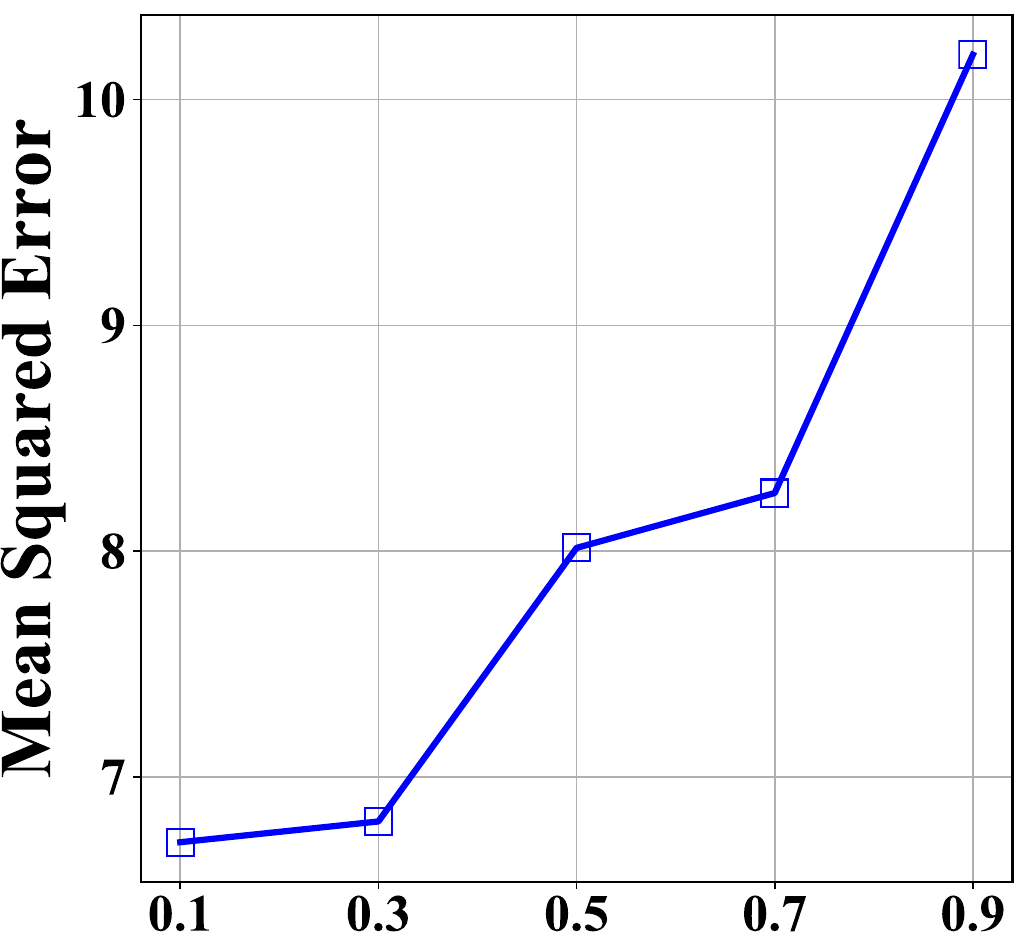}}
	\caption{Influence on parameters.}
	\label{drop}
\end{figure}

\paragraph{Parameter sensitivity analysis.}
We analyse two important parameters, i.e., the probability of masking features and the probability of removing edges in generating two views during graph augmentation. We carry out experiments on the Aids700nef dataset with different parameters as an example, and Fig. \ref{drop} shows the effect of parameter sensitiveness on the CGMN performance. The horizontal coordinates represent the probability of masking node features or removing edges, and the vertical coordinates represent the MSE ($10^{-3}$). For example, $0.1$ means randomly masking node features with probability $10\%$. In Fig. \ref{drop} (a) and (b), despite the fluctuations in the line chart, the overall trend of MSE values is increasing. This means that increasing the modification probability excessively may distort the graph structures and features information, resulting in significant performance degradation.

\section{Conclusion}
Graph similarity learning is pivotal in graph problems. In this paper, we propose a novel unsupervised graph similarity learning method CGMN. Using a siamese GNN as the backbone, we design a cross-view interaction strategy to enrich the self-supervised signal. To further match nodes across graphs, we introduce a cross-graph interaction mechanism to conduct cross-graph matching between nodes. The experimental results demonstrate the superiority and effectiveness of the new approach compared with the SOTA methods.

\section*{Acknowledgments}
This work was partly supported by the National Natural Science Foundation of China under grants 61876128 and Meituan Project. 
\bibliographystyle{named}
\bibliography{ijcai22}

\end{document}